\def\year{2020}\relax
\documentclass[letterpaper]{article} 
\usepackage{aaai20}  
\usepackage{times}  
\usepackage{helvet} 
\usepackage{courier}  
\usepackage[hyphens]{url}  
\usepackage{graphicx} 
\usepackage{graphicx}  
\usepackage{subfigure} 
\usepackage{amsmath}
\usepackage{amssymb}
\usepackage{amsthm}
\usepackage{amsopn}
\usepackage{amsmath,bm}
\usepackage{array}
\usepackage{booktabs}
\usepackage{algorithm}
\usepackage{algorithmic}
\usepackage{enumerate}
\usepackage{color}
\usepackage{multirow}
\usepackage{adjustbox}

\newcommand{\ie}{\emph{i.e. }}

\title{Efficient Residual Dense Block Search for Image Super-Resolution}
\author{Dehua Song$^{1}$, Chang Xu$^{3}$, Xu Jia$^{1}$\thanks{The author should be considered as equal second author.}, Yiyi Chen$^{2}$, Chunjing Xu$^{1}$, Yunhe Wang$^{1}$ \\
    \normalsize$^1$ Huawei Noah's Ark Lab. \ \ \normalsize$^2$ Huawei CBG. \\
    \normalsize$^3$ School of Computer Science, Faculty of Engineering, The University of Sydney.\\
    \small\texttt{\{dehua.song, x.jia, yiyi.chen, xuchunjing, yunhe.wang\}@huawei.com;}\\
    \small\texttt{c.xu@sydney.edu.au}
}

\begin{document}

\maketitle

\begin{abstract}

Although remarkable progress has been made on single image super-resolution due to the revival of deep convolutional neural networks, deep learning methods are confronted with the challenges of computation and memory consumption in practice, especially for mobile devices. Focusing on this issue, we propose an efficient residual dense block search algorithm with multiple objectives to hunt for fast, lightweight and accurate networks for image super-resolution. Firstly, to accelerate super-resolution network, we exploit the variation of feature scale adequately with the proposed efficient residual dense blocks. In the proposed evolutionary algorithm, the locations of pooling and upsampling operator are searched automatically. Secondly, network architecture is evolved with the guidance of block credits to acquire accurate super-resolution network. The block credit reflects the effect of current block and is earned during model evaluation process. It guides the evolution by weighing the sampling probability of mutation to favor admirable blocks. Extensive experimental results demonstrate the effectiveness of the proposed searching method and the found efficient super-resolution models achieve better performance than the state-of-the-art methods with limited number of parameters and FLOPs. 
 
\end{abstract}

\section{Introduction}

Single image super-resolution (SISR) is to generate a high-resolution image from its degraded low-resolution version. It has broad applications in photo editing, surveillance and medical imaging. SISR is hard due to the fact that reconstructing high resolution image from a low resolution image is a many-to-one mapping. To address this ill-posed inverse problem, various methods were introduced like interpolation based methods, dictionary-based methods and deep learning based methods.

In recent years, convolutional neural network-based (CNN-based) super-resolution methods have been extensively studied and have achieved tremendous improvement in terms of peak signal-to-noise (PSNR). From the pioneering SRCNN \cite{dong2014learning} to the most current RDN 
\cite{zhang2018residual}, the overall performance of image super-resolution has dramatically boosted, and with it come the increases of the number of parameters and the amount of computation. For example, RDN contains $22M$ parameters and requires $10,192 G$ FLOPs (floating point operations). This severely restricts CNN-based super-resolution methods to be deployed on mobile devices that possess limited computing and storage resources. It is therefore urgent and important to develop fast, lightweight and accurate super-resolution networks for real-world applications.

A natural idea to address the aforementioned challenges is designing efficient operator or block in the networks \cite{zhang2018shufflenet,wang2018learning,wang2018packing}. Recursive operator has been widely investigated to reduce redundant parameters of super-resolution network, such as DRCN \cite{kim2016deeply}, DRRN \cite{tai2017image}, and MemNet \cite{tai2017memnet}. In addition, researchers have tried to construct efficient super-resolution blocks with squeeze operations and group operations \cite{ahn2018fast,shi2016real}. However, blocks in a hand-crafted super-resolution network usually share the same architecture, which might not always be an optimal solution for the task, as the flexibility of the block and capacity of the whole network could be decreased. 

Given proper search space, neural architecture search (NAS) is helpful in deriving the optimal architectures for different tasks. Recently, this automation has led to state of the art accuracy on classification problems~\cite{zoph2018learning,real2019regularized,liu2018darts}. This success has been reproduced in many other computer vision tasks, e.g. segmentation \cite{liu2019auto} and image restoration \cite{suganuma2018exploiting}. NAS has rarely been investigated for lightweight image super-resolution. The most relevant work here may be the MoreMNAS~\cite{chu2019fast}, which utilized multi-objective NAS to pursue the tradeoff between restoration and simplicity of super-resolution models. However, the efficiency of network is imposed strict restriction by the large amounts of computation of the whole nonlinear mapping process, which is computed on the same scale entirely. In addition, it searches architecture on inner cell level and outer network level simultaneously with basic operators. It is time-consuming and harmful to mobile devices due to the non-regularized combinations of arbitrary basic operators~\cite{ma2018shufflenet}.

\begin{figure*}[tp]
\begin{center}
\scalebox{0.98}{
\includegraphics[width=0.98\textwidth]{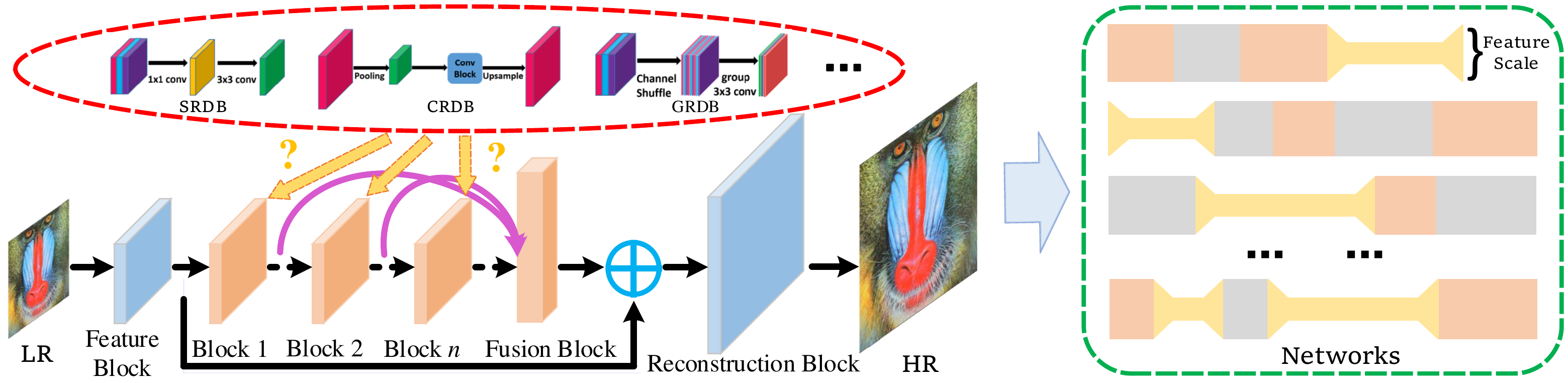}
}
\end{center}
\vspace{-1.5em}
\caption{The diagram of super-resolution neural architecture search. It employs the proposed efficient residual dense blocks to exploit the variation of feature scale adequately for efficient super-resolution network.}
\label{Fig:flowchart}
\vspace{-1.0em}
\end{figure*}

In this paper, we propose the efficient residual dense block search algorithm for image super-resolution. Pooling operation is a double-edged sword in image super resolution. It often leads to the information loss but will reduce the computation in the subsequent process and realize acceleration. Instead of eschewing the pooling as classical methods, we integrate local residual learning designed in a contextual residual dense block and global feature fusion to fix the defect of the pooling. In addition, shrink residual dense block and group residual dense block are developed to further reduce parameters. Given these three types of building blocks, we resort to evolutionary algorithm to search for the optimal network architecture for image super-resolution. Block credit is introduced to measure the effectiveness of a block, and guides the mutation by weighing the sampling probability of mutation to favor admirable block. The proposed method exploits valuable information from model evaluation process to guide the network evolution for effective super-resolution network and convergence acceleration. Experimental results demonstrate the effectiveness of the proposed searching method and the found efficient super-resolution models achieve better performance than the state-of-the-art methods with limited number of parameters and FLOPs.

\section{Related Works}

Deep learning based super-resolution methods have achieved significant promotion and been investigated extensively. \cite{dong2014learning} firstly employed convolutional neural network with three layers into image super-resolution task and made a noticeable progress. After that, researchers explored deeper and deeper network with the assistance of shortcut operator and dense connection, like VDSR \cite{kim2016accurate}, RAAN~\cite{xin2019residual}, EDSR~\cite{lim2017enhanced} and RDN~\cite{zhang2018residual}. Another direction is designing lightweight and fast super-resolution for practical applications. FSRCNN~\cite{dong2016accelerating} and ESPCN~\cite{shi2016real} delayed the position of upsampling operator to accelerate   super-resolution network. Recursive operator is widely employed in super-resolution to reduce parameters of network \cite{tai2017image,tai2017memnet}. Besides, several works~\cite{ahn2018fast,hui2018fast} explored light-weight and fast super-resolution model with squeeze operation and group convolution. Recently, researchers attempted to apply NAS to super-resolution task \cite{chu2019fast}.

\section{Super-resolution Neural Architecture Search}

To acquire fast, lightweight and accurate super-resolution networks, evolution-based NAS algorithm are employed and adapted to super-resolution task. The  diagram of the proposed method is illustrated in Fig. \ref{Fig:flowchart}.

\subsection{Efficient Residual Dense Blocks}

Proper search space is crucial to the performance of NAS algorithm. The success of NAS method based on meta-architectures (e.g. block/cell) attracts us to extend it to super-resolution architecture search, which is efficient and effective in many tasks~\cite{tan2018mnasnet,guo2019single}. Besides, blocks are widely used and investigated in super-resolution task. Recently, RDN~\cite{zhang2018residual} has exhibited the power of residual dense block (RDB). However, weights and computation of block grows rapidly along with the increasing of convolution number due to the dense concatenation. Searching for fast and lightweight network, lean block is essential. Considering this issue, this paper proposes three lean residual dense blocks, which reduce the parameter and computation of block in three aspects: channel number, convolution filter and feature scale.

\begin{figure*}
\centering
\scalebox{0.94}{
\includegraphics[width=0.94\textwidth]{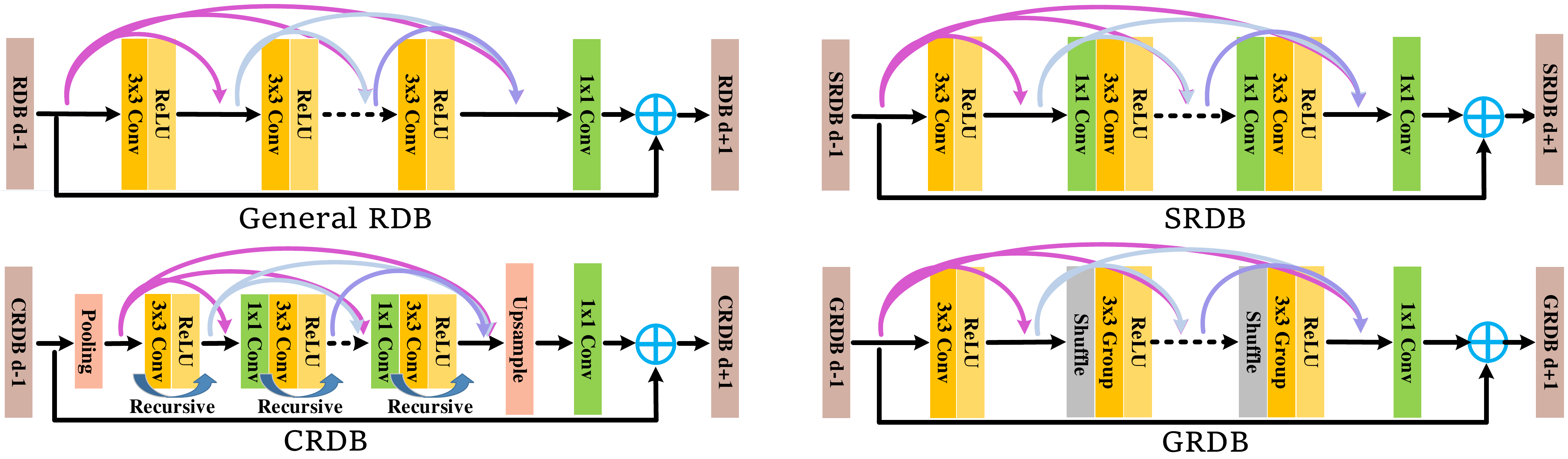}
}
\vspace{-0.8em}
\caption{The architecture of proposed efficient residual dense block (SRDB, GRDB, CRDB).}
\label{Fig:CRDB}
\vspace{-1.0em}
\end{figure*}

\textit{\textbf{Shrink Residual Dense Block (SRDB)}}  Massive parameters and computations of block are mainly because of swift growth of channel number. An intuitive idea to reduce the parameter and computation is reducing the channel number. Here, we employ $1 \times 1$ convolution to squeeze the channels of feature maps to construct a lean block. The architecture of SRDB is illustrated in Fig.~\ref{Fig:CRDB}.

\textit{\textbf{Group Residual Dense Block (GRDB)}} Another way is reducing the parameters of convolution filter. Group convolution \cite{zhang2018shufflenet} is employed to construct lean block. Since the input of convolution concatenates multiple preceding feature maps, if we directly utilize the group convolution in block, a single output channel cannot see all preceding layers' features. Therefore, the channel shuffle strategy \cite{zhang2018shufflenet} is utilized before group convolution to alleviate above issue. The architecture of GRDB is illustrated in Fig.~\ref{Fig:CRDB}.

\textit{\textbf{Contextual Residual Dense Block (CRDB)}} The third way to depress the computation of block is reducing the scale of feature. Different from the classification task, pooling operator is seldom used in super-resolution networks. It's mainly because pooling operation ignores information and might deteriorate the performance of super-resolution without elegant design. To alleviate this issue, we adopt local residual connection and block feature concatenation to transit original information for image reconstruction, see Fig.~\ref{Fig:flowchart}. In addition to reducing computation, pooling operator also expands the receptive field to obtain more context information, which is conducive to reconstruction high-resolution image. Besides, recursive operator~\cite{tai2017image} is employed to further amplify the receptive field. Hence, the proposed block is named as Contextual Residual Dense Block (CRDB). The architecture of CRDB is illustrated in Fig.~\ref{Fig:CRDB}. CRDB contains four components: pooling, recursive convolutional layer, upsample operator and local residual operator. Sub-pixel convolution \cite{shi2016real} is used to upsample the feature maps for feature fusion due to its efficiency.

\subsubsection{Efficiency Analysis} 

The effect of group convolution of GRDB is analyzed deeply in literature \cite{ahn2018fast}. Here, we place emphasis on the effect of CRDB, which contains three advantages: reducing computation, expanding receptive field and decoupling FLOPs and parameter. Firstly, pooling operator reduces FLOPs by decreasing the size of feature maps. Let $K$ be the kernel size and $C$ be the input convolution number of CRDB. Both the width and height of feature maps of general RDB are $S$ and the growth rate (output channel number of each $K \times K$ convolution layer) is $G_r$. For simple, the input and output channel number of CRDB is also $G_r$. The recursive number is $R$. Then, the FLOPs of general RDB is:
\begin{equation}
\label{FLOPS}
FLOPs_{rdb} = 2G_r^2S^2(CK^2 + C(C+3)/2),
\end{equation}
The FLOPs of CRDB is:
\begin{equation}
\label{FLOPS_crdb}
FLOPs_{crdb} = 2G_r^2S^2(CK^2R/4 + (C+7)(C+1)/8),
\end{equation}
If $R = 1$, then the $FLOPs_{crdb} = FLOPs_{rdb}/4 + (5C+7)G_r^2S^2/4 \approx FLOPs_{rdb}/4$. It reveals that pooling operator approximately reduces 75\% FLOPs of RDB. If $R = 4$, the $FLOPs_{crdb}$ is still less than $FLOPs_{rdb}$. Secondly, both pooling operator and recursive operator could expand the receptive field. Finally, without pooling and recursive operator, the FLOPs and parameters of network is linearly correlated: $FLOPs = S^2 \times Param$, which is also verified in literature \cite{chu2019fast}. In contrast, our method allows evolutionary algorithm search the architecture space with three independent objectives: PSNR, FLOPs and parameter.

\subsection{Efficient Residual Dense Block Search}

Evolutionary algorithm is employed to search the neural architecture because of its excellent performance~\cite{real2019regularized}. Neural architecture search is confronted with the challenge of high computational consumption. In contrast with classification, this issue is more serious in super-resolution neural architecture search. Because feature maps are computed in the same scale or high resolution scale, which consumes more computation in model evaluation procedure. It requires us to employ the characteristics of super-resolution to alleviate this issue.

\subsubsection{Search Space}

Super-resolution neural network is usually composed of three stages: feature extraction, nonlinear mapping and reconstruction \cite{dong2014learning}. This work mainly focuses on searching the architecture of backbone (nonlinear mapping) which is critical to high-quality super-resolution model, see Fig.~\ref{Fig:flowchart}. The block number and architecture of each block in backbone are searched automatically. The architecture contains block type and hyperparameter (i.e. number of convolutional layers, channel number of convolutional layer, recursive number, output channel number of block). To accelerate architecture searching, the type of block is quantized as the proposed three lean blocks: SRDB, GRDB and CRDB. Searching the type and hyperparameter of block could choose the location and convolution number of low resolution features automatically, which is conducive to exploiting the redundant computation of feature scale and accelerating super-resolution network. Each block is encoded as $\{block \ state, block \ type, hyperparameter\}$, which is corresponding to the whole search space (block number, block type and the hyperparameter of block). The detail of search space is introduced in section \textit{experiments}.

\subsubsection{Evolutionary Algorithm}

The model training procedure possesses abundant information which is conducive to facilitating the evolution. It attracts us to exploit the information of learning procedure and investigate how to guide the evolution. Inspired by intermediate supervision in the middle of architecture \cite{xu2017fast,lai2017deep}, we employ joint loss to acquire the evaluation of the whole network and each block simultaneously during model evaluation procedure. On this basis, we propose a heuristic approach to guide its mutation operation by taking into account appearance of genes (choice blocks) in the evolved chromosome (architectures). The whole algorithm is shown in Algorithm \ref{alg:evolution}.

The evolutionary algorithm maintains a population of models $H$ in which each individual $h$ is initialized by randomly selecting a choice block in uniform distribution for each position. To evolve the population, each individual is trained with training set and tested on validation set to compute fitness. In super-resolution, the fitness of model is measured by PSNR between high resolution images and super-resolution images on the validation set. After this, we select the top $T$ elites as elitism according to the fitness of population and evolve it in cycles. The next generation are generated with two operation: crossover and guided mutation. The parents of crossover are selected as the strategy of roulette wheel selection.

\begin{algorithm}[H]
\caption{Guided evolutionary algorithm}
\label{alg:evolution}
\begin{algorithmic}
\REQUIRE number of generations $G$, population size $\lambda $, mutation probability $r$, elitism number $T$, Training Set $D_{train}$, Validation Set $D_{val}$.
\STATE \textbf{Initialization:} (i) population $H$, (ii) block credit matrix $M_b$, (iii) fitnesses $F_H$ of all individuals in population.
\WHILE {$g < G$}
\STATE elitism $E_g$ $\leftarrow$ $select$($\{ H, E_{g-1}\}$, $F_H$, $T$)
\FOR{$i=1$ to $\frac{\lambda}{2}$}
\STATE ${children}_i$ $\leftarrow$ $GuidedMutate$($E_g, r, M_b$)
\STATE ${children}_{\frac{\lambda}{2}+i}$ $\leftarrow$ $Crossover$( $\{ H, E_{g-1}\}$ )
\ENDFOR
\STATE $H$ $\leftarrow$ $children$
\STATE $models$ $\leftarrow$ $Train$($H$, $D_{train}$)
\STATE $F_H, M_b'$ $\leftarrow$ $Evaluate$($models$, $D_{val}$)
\STATE $M_b$ $\leftarrow$ $update$($M_b$, $M_b'$) as Eq.~\ref{mat_update}
\STATE $g = g + 1$
\ENDWHILE
\ENSURE elitism $E_g$
\end{algorithmic}
\end{algorithm}

\textit{\textbf{Block Credit}} The credit of a block is defined as performance gain by adding it on top of preceding blocks, which demonstrates its effectiveness. The credit of block is defined as follows:
\vspace{-0.15em}
\begin{equation}
\label{Node_cre_def}
c = f_{add} - f_{bef},
\end{equation}
where $f_{add}$ and $f_{bef}$ denote the fitness (PSNR) of network with current block and without current block on the end of network, respectively.
It is obvious that block credit is related to current block and the preceding blocks. Hence, the block credit is redefined as $c_m = \frac{1}{N}\sum_{i=1}^{N} {c_i}$, where $c_i$ denotes block credit with different combination of preceding blocks and $N$ denotes the number of combination. The block credits of different block or different location (depth) are different, hence the block credit matrix $M_b$ is composed of different block credits $c_m(j,l)$, where $j$ denotes block index and $l$ denotes the index of depth.

To acquire the block credit and model fitness simultaneously during individual performance evaluation procedure, we employ an adaptive joint loss combining all intermediate loss. It is defined as: $\mathcal L = \sum_{l=1}^{L} {\eta}_l(t) \mathcal L_l$, where $\mathcal L_l $ denotes the $L1$ loss between the reconstructed super-resolution images of block $l$ and high-resolution images. $\eta$ is the weight coefficient of each block and $t$ is the training epoch which controls the distribution of blocks' coefficients. Computing all combinations' fitness of preceding blocks is time consuming, hence the block credit are computed approximately and updated constantly during evolution procedure. During the evolution procedure, we update the block credit matrix $M_b$ as follows:
\vspace{-0.2em}
\begin{equation}
\label{mat_update}
{c_m^{\prime}{(j,l)}} = \alpha{c_m{(j,l)}} + (1- \alpha){c}{(j,l)},
\end{equation}
where $\alpha$ is the update coefficient of block fitness.

\textit{\textbf{Guided Mutation}} In general mutation, it is common to choose a block architecture randomly instead of old architecture. It is a good idea when there is no other information about block except architecture, although the mutation has no direction on the whole evolution procedure which results in time-consuming search. However, we acquire the block credits during model evaluation procedure, which can be used to guide the mutation to accelerate searching and find better architecture. Inspired by parent selection strategy of crossover, we propose a novel guided mutation: the architecture of mutated block is selected by typical proportionate roulette wheel method according to block credits. This strategy favors the block architecture with higher credit in search space, which is conducive to finding better architecture. To enhance the diversity of different block, the block credit is normalized as follows:
\vspace{-0.5em}
\begin{equation}
\label{Node_fit_norm}
{c_n}{(j,l)} = c_m{(j,l)} - (\min_l(c_m{(j,l)}) - \epsilon),
\vspace{-0.5em}
\end{equation}
where $\epsilon$ is a tiny constant and $c_n$ denotes normalized block credit. If the block is in layer $l_b$, then the probability that block $b_j$ is selected is $p$, which is defined as follows:
\vspace{-0.5em}
\begin{equation}
\label{cell_prob_sim}
p_{select}(b_j|l=l_b) = \frac{c_n{(j,l_b)}}{\sum_{j=1}^{N_b}{c_n}{(j, l_b)}},
\vspace{-0.5em}
\end{equation}
where $N_b$ denotes the number of block architecture in search space.

\subsubsection{Performance Evaluation}

For single image super-resolution, the performance of model is usual evaluated by PSNR and SSIM (structural similarity index) in the literature. During the evolution, PSNR is adopted to evaluate the accuracy of searched architecture. To search for lightweight and fast model, we evaluate the parameter number of neural network and the computational FLOPs. It is a typical multi-objective evolution problem. Different from MoreMNAS~\cite{chu2019fast}, we add pooling and recursive operator in the search space, which decouples the FLOPs and parameter number. Our method offers lean networks with different proportions of FLOPs and parameter. For example, mobile phone possesses comparatively sufficient memory, but requires short runtime.

Solving multi-objective evolution problem, two strategies can be employed: NSGA-II method ~\cite{chu2019fast} and converting the multi-objective evolution to a single objective problem. NSGA-II method offers the whole Pareto optimality and the objective function is defined as:
\vspace{-0.3em}
\begin{equation}
\label{NAGA_obj}
\begin{split}
\max_{net \in Q} \ objs(net) = & \{ psnr(net), -param(net), \\
& -flops(net) | net \in Q \},
\end{split}
\end{equation}
Sometimes, we may only need the best model with the constraint on FLOPs and parameters. Hence, we can constrain the FLOPs and parameters, and maximize the PSNR of model, which is defined as follows:
\vspace{-0.2em}
\begin{equation}
\label{single_obj} 
\begin{split}
& \max_{net \in Q} \ psnr(net) \\
& s.t. \ \ param(net) < W_{net}, \ \ flops(net) < V_{net}
\end{split}
\end{equation}
where $W_{net}$ and $V_{net}$ are the upper constraints of parameter and FLOPs, respectively. 

\section{Experiments}
\label{headings}

\subsection{Datasets and Implementation}
\label{experiment_detail}

\textit{\textbf{Datasets}} Most widely used dataset DIV2K \cite{timofte2017ntire} is adopted in this paper. It consists of 800 training images and 100 validation images. During the evolution and retrain procedures, the SR models are trained with DIV2K training set and evaluated with validation set. To test the performance of the model, we employ four benchmark datasets: Set5, Set14, B100 and Urban100. The super-resolution results are evaluated with PSNR and SSIM on Y channel of YCbCr space.

\textit{\textbf{Searching Detail}} For the proposed evolutionary algorithm, the detail search space is: block type $\{S, G, C\}$, convolutional layer number $\{4,6,8\}$, growth rate and output channel number $\{ 16, 24, 32, 48, 64\}$, recursive number $\{1, 2, 3, 4\}$. The recursive number ${1}$ denotes a normal convolution layer. The number of children is 16 which is composed of 8 children mutated from elitism and 8 children crossovered with parents. The number of generation $G$ is 40 and the mutation probability is $0.2$. The coefficient $\alpha$ is $0.9$ and constant $\epsilon$ is $0.001$. Coefficient $\eta(t)$ is updated every 10 epochs. ${\eta}_L$ is initialized with $0.0625$ and multiplied 2 every period. All of the other coefficients ${\eta}_l$ are $ \frac{1-{\eta}_L}{L-1}$. To enhance the difference of block credits, we employ the square of block credit to guide the mutation, ~\ie $p_{select}(b_j) = c_n^2{(j)}/{\sum_{j=1}^{N_b}{c_n^2}{(j)}}$. For the phenotype, the maximum block number is 20 and the minimum active block number is 5. During evolution, we crop $32 \times 32$ RGB patches from LR image as input for training. We train each model for 60 epoch with a mini-batch of size 16. The evolution procedure is performed on single Tesla V100 with 8 GPUs and spends about one day.

\textit{\textbf{Retraining Detail}} After the evolution stage, the selected super-resolution model is retrained on DIV2K training set. We put $64 \times 64$ RGB patches of LR images into network for training. The training data augmented with random crop, horizontal flips and 90 degree rotation. Our models are trained with ADAM optimizer with setting $\beta_1 = 0.9$, $\beta_2 = 0.999$, and the initialized learning rate is $10^{-4}$. The learning rate decreases half for every 300 epochs during the whole 1000 training epoch. 

\begin{table}[t]
\vspace{-1.5mm}
\begin{center}
\caption{Ablation investigation of the proposed efficient super-resolution blocks ('O' denotes no recursive operator, 'L','C' and 'F' denotes the arrangement of CRDB ).}
\label{pool_ablation}
\setlength{\tabcolsep}{1.25mm}{
\begin{tabular}{p{1.7cm}<{\centering}|p{1.1cm}<{\centering}|p{0.95cm}<{\centering}| p{0.90cm}<{\centering} | p{0.90cm}<{\centering} | p{1.28cm}<{\centering}}
\hline
\multirow{2}*{Model} & Multi- & Params & Set14 & B100 & Urban100 \\
~ &  Adds(G) &  (K) &  PSNR & PSNR & PSNR \\
\hline\hline
RDN (base) & 235.6 & 1017 & 33.44 & 32.02 & 31.94\\
\hline
GRDN & 235.6 & 1017 & 33.55 & 32.12 & 32.12\\
\hline
SRDN & 235.8 & 1019 & 33.54 & 32.13 & 32.15\\
\hline
CRDN-O-L & 163.9 & 991 & 33.46 & 32.08 & 31.81\\
\hline
CRDN-O-C & 163.9 & 991 & 33.50 & 32.09 & 31.86\\
\hline
CRDN-O-F & 163.9 & 991 & 33.52 & 32.11 & 31.90\\
\hline
CRDN & 225.2 & 1018 & 33.57 & 32.15 & 32.20\\
\hline
ESRN & 226.8 & 1014 & \textbf{33.71} & \textbf{32.23} & \textbf{32.37}\\
\hline
\end{tabular}}
\end{center}
\vspace{-5mm}
\end{table}

\begin{table}[t]
\begin{center}
\caption{Comparison experiments on general mutation and guided mutation ($\times 2$ scale super-resolution).}
\label{abl_guided}
\setlength{\tabcolsep}{1.35mm}{
\begin{tabular}{p{1.25cm}|p{1.12cm}<{\centering}|p{1.0cm}<{\centering}| p{1.0cm}<{\centering} | p{1.0cm}<{\centering} | p{1.33cm}<{\centering}}
\hline
\multirow{2}*{Mutation} & Multi- & Params & Set14 & B100 & Urban100 \\
~ &  Adds(G) &  (K) &  PSNR & PSNR & PSNR \\
\hline\hline
General & 228.4 & 1018 & 33.65 & 32.18 & 32.25\\
\hline
Guided & 226.8 & 1014 & \textbf{33.71} & \textbf{32.23} & \textbf{32.37}\\
\hline
\end{tabular}}
\end{center}
\vspace{-2.0em}
\end{table}

\subsection{Ablation Study of Efficient Block}

Extensive ablation experiments are designed to evaluate the effectiveness of the proposed three efficient super-resolution blocks. Firstly, we construct a standard light RDN with 1017K parameter as the basic model, which contains 4 RDBs with six convolution layers of 32 channels (growth rate). Then, we use single type of proposed block to construct super-resolution model without architecture search and evaluate their performance. In order to be consistent with previous super-resolution literatures~\cite{ahn2018fast,chu2019fast}, Multi-Adds is employed to evaluate the computation of models and it is computed with 720p high-resolution image (~\ie $1280 \times 720$). Experimental results summarized in Table~\ref{pool_ablation} shows that our efficient blocks achieve better performance than baseline with the constraint of Multi-Adds and parameters.

\begin{figure}
\vspace{-0.5em}
\centering
\centering
\includegraphics[width=0.46\textwidth]{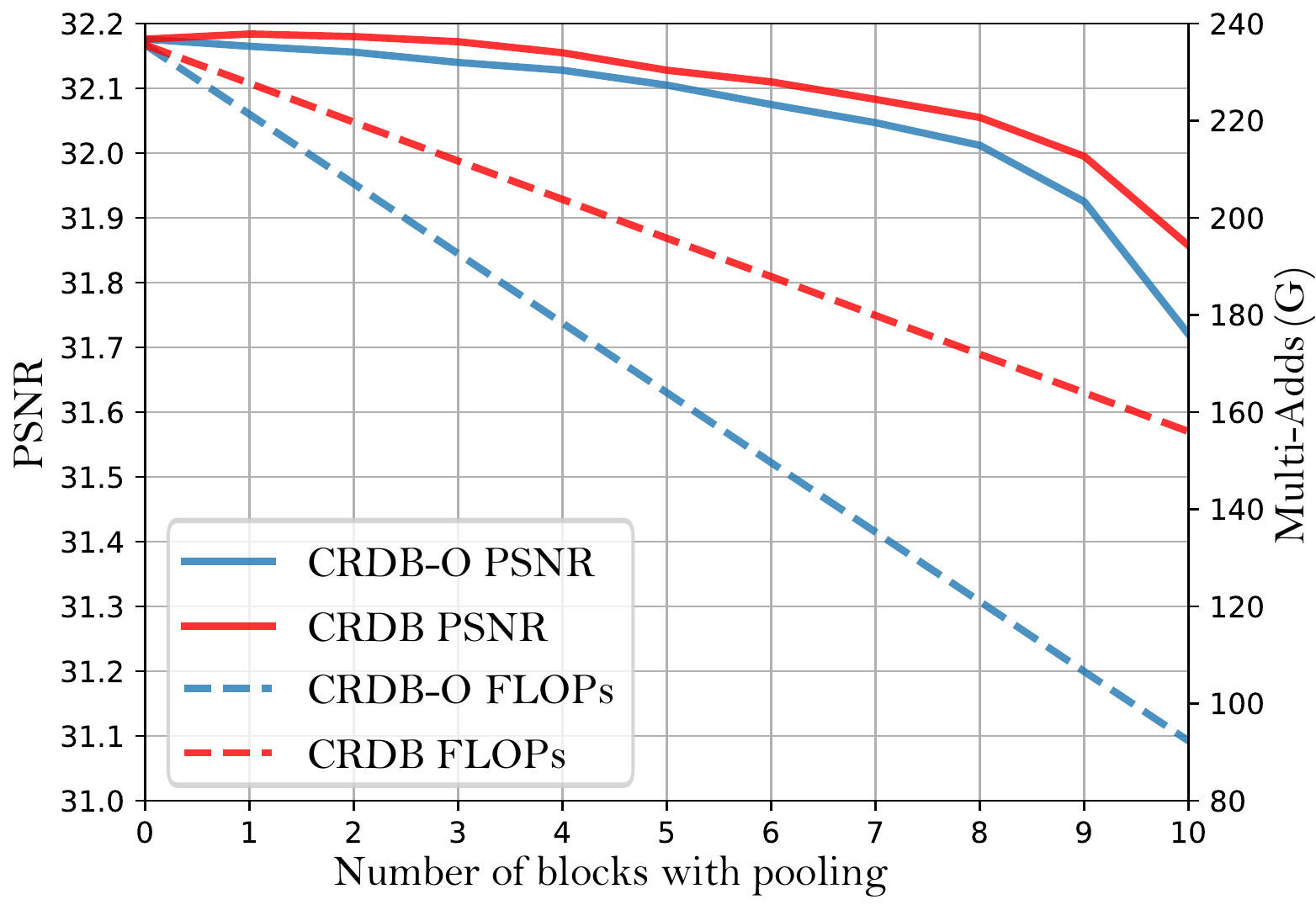}
\vspace{-1.2em}
\caption{Performance of super-resolution model with different number of pooling blocks ('O' denotes no recursive operator in CRDB).}
\label{pool_diff}
\end{figure}

\begin{figure}
\centering
\includegraphics[width=0.45\textwidth]{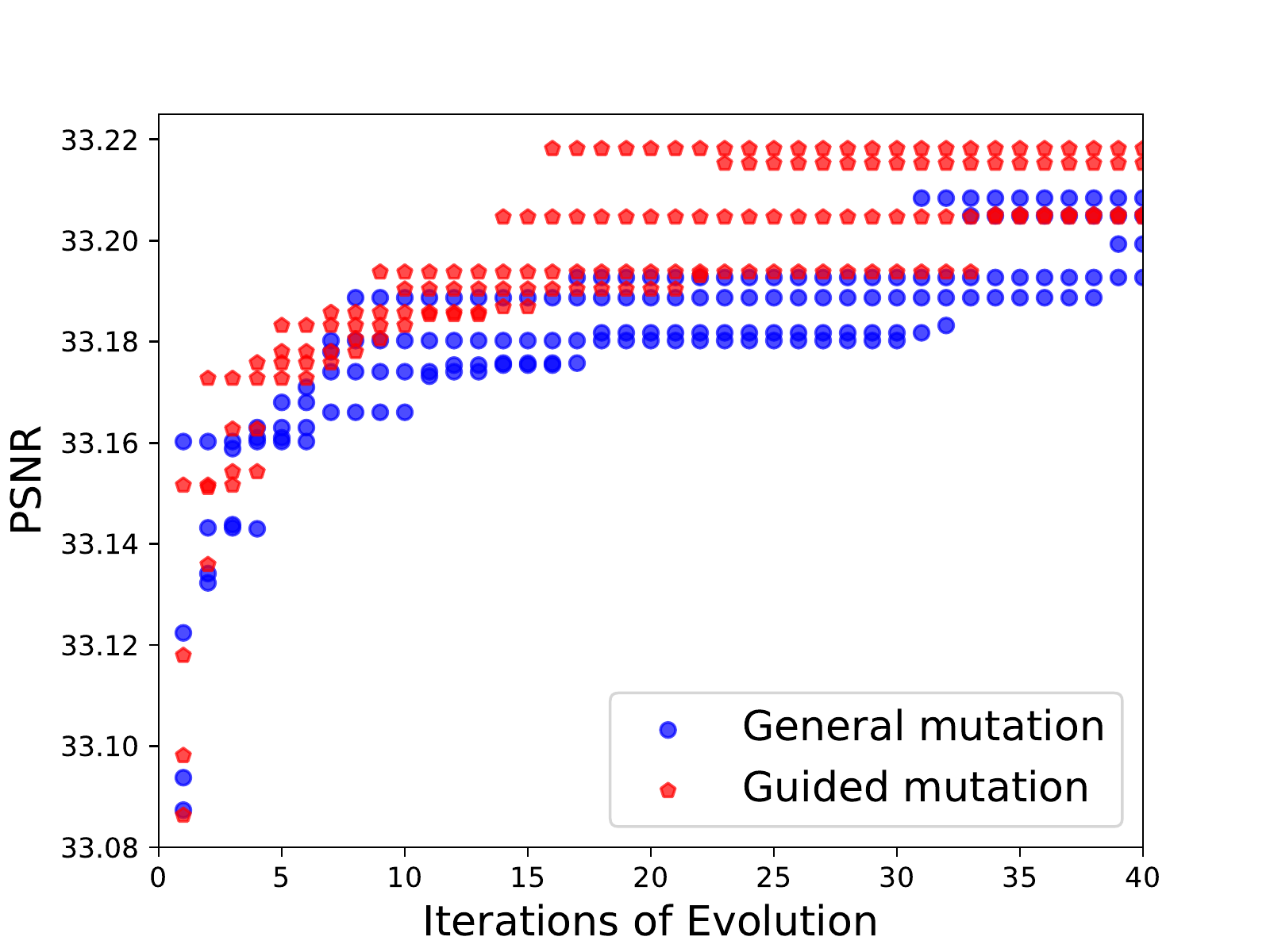}
\vspace{-1.2em}
\caption{Evolutionary architecture search with different mutation strategies.}
\label{mutate_diff}
\vspace{-0.5em}
\end{figure}

\begin{table*}[t]
\begin{center}
\caption{Quantitative results of the state-of-the-art super-resolution models on $\times 2$ scale super-resolution task (the best results are emphasized with bold).}
\label{compare_sota}
\setlength{\tabcolsep}{1.40mm}{
\begin{tabular}{@{}  c|c|p{1.7cm}<{\centering} p{1.3cm}<{\centering}|cccc}
\hline
\multirow{2}*{Type} & \multirow{2}*{Model} & Multi-Adds & Parameter &  Set5 & Set14 & B100 & Urban100 \\
~  & ~ & (G) &  (K) &  PSNR/SSIM & PSNR/SSIM & PSNR/SSIM & PSNR/SSIM \\
\hline\hline
\multirow{4}*{Slow} & VDSR (Kim et al. 2016a) & 612.6 & 665 & 37.53/0.9587 & 33.03/0.9124 & 31.90/0.8960 & 30.76/0.9140\\
~ & DRCN (Kim et al. 2016b) & 17974.3 & 1774 & 37.63/0.9588 & 33.04/0.9118 & 31.85/0.8942 & 30.75/0.9133\\
~ & MemNet~\cite{tai2017memnet} & 2662.4 & 677 & 37.78/0.9597 & 33.28/0.9142 & 32.08/0.8978 & 31.51/0.9312\\
~ & RDN~\cite{zhang2018residual} & 5096.2 & 22114 & 38.24/0.9614 & 34.01/0.9212 & 32.34/0.9017 & 32.89/0.9353\\
\hline\hline
\multirow{4}*{Fast} & CARN (Anh et al. 2018) & 222.8 & 1592 & 37.76/0.9590 & 33.52/0.9166 & 32.09/0.8978 & 31.92/0.9256\\
~ & FALSR-A~\cite{chu2019fast} & 234.7 & 1021 & 37.82/0.9595 & 33.55/0.9168 & 32.12/0.8987 & 31.93/0.9256\\
~ & ESRN (ours)& 228.4 & \textbf{1014} & \textbf{38.04/0.9607} & \textbf{33.71/0.9185} & \textbf{32.23/0.9005} & \textbf{32.37/0.9310}\\
~ & ESRN-F (ours)& \textbf{128.5} & 1019 & 37.93/0.9602 & 33.56/0.9171 & 32.16/0.8996 & 31.99/0.9276\\
\hline\hline
~ & SRCNN~\cite{dong2014learning} & 52.7 & 57 & 36.66/0.9542 & 32.42/0.9063 & 31.36/0.8879 & 29.50/0.8946 \\
Very & CARN-M~(Ahn et al. 2018) & 91.2 & 412 & 37.53/0.9583 & 33.26/0.9141 & 31.92/0.8960 & 31.23/0.9144\\
Fast & FALSR-B~\cite{chu2019fast} & 74.7 & 326 & 37.61/0.9585 & 33.29/0.9143 & 31.97/0.8967 & 31.28/0.9191\\
~ & ESRN-V (ours) & \textbf{73.4} & \textbf{324} & \textbf{37.85/0.9600} & \textbf{33.42/0.9161} & \textbf{32.10/0.8987} & \textbf{31.79/0.9248}\\
\hline
\end{tabular}}
\end{center}
\vspace{-1.2em}
\end{table*}

\begin{table*}[ht]
\begin{center}
\caption{Quantitative results of the state-of-the-art super-resolution models on $\times 3$ and $\times 4$ scale.}
\label{scale_sota}
\setlength{\tabcolsep}{1.4mm}{
\begin{tabular}{@{}  c|c|p{1.7cm}<{\centering} p{1.3cm}<{\centering}|cccc}
\hline
\multirow{2}*{Scale} & \multirow{2}*{Model} & Multi-Adds & Parameter &  Set5 & Set14 & B100 & Urban100 \\
~  & ~ &  (G) &  (K) &  PSNR/SSIM & PSNR/SSIM & PSNR/SSIM & PSNR/SSIM \\
\hline\hline
\multirow{10}*{$\times 3$} & SRCNN~\cite{dong2014learning} & 52.7 & 57 & 32.75/0.9090 & 29.28/0.8209 & 28.41/0.7863 & 26.24/0.7989 \\
~ & VDSR~(Kim et al. 2016a) & 612.6 & 665 & 33.66/0.9213 & 29.77/0.8314 & 28.82/0.7976 & 27.14/0.8279\\
~ & MemNet~\cite{tai2017memnet} & 2662.4 & 677 & 34.09/0.9248 & 30.00/0.8350 & 28.96/0.8001 & 27.56/0.8376\\
~ & SelNet~(Choi et al. 2017) & 120.0 & 1159 & 34.27/0.9257 & 30.30/0.8399 & 28.97/0.8025 & - \\
~ & CARN~(Ahn et al. 2018) & 118.8 & 1592 & 34.29/0.9255 & 30.29/0.8407 & 29.06/0.8034 & 28.06/0.8493\\
~ & CARN-M~(Ahn et al. 2018) & 46.1 & 412 & 33.99/0.9236  &30.08/0.8367 & 28.91/0.8000 & 27.55/0.8385\\
~ & ESRN (ours)& 115.6 & 1014 & \textbf{34.46/0.9281} & \textbf{30.43/0.8439} & \textbf{29.15/0.8072} & \textbf{28.42/0.8579}\\
~ & ESRN-F (ours)& 71.7 & 1019 & 34.32/0.9268 & 30.35/0.8410 & 29.09/0.8046 & 28.11/0.8512\\
~ & ESRN-V (ours)& \textbf{36.2} & \textbf{324} & 34.23/0.9262 & 30.27/0.8400 & 29.03/0.8039 & 27.95/0.8481\\
\hline\hline
\multirow{10}*{$\times 4$} & SRCNN~\cite{dong2014learning} & 52.7 & 57 & 30.48/0.8628 & 27.49/0.7503 & 26.90/0.7101 & 24.52/0.7221 \\
~ & VDSR~(Kim et al. 2016a) & 612.6 & 665 & 31.35/0.8838 & 28.01/0.7674 & 27.29/0.7251 & 25.18/0.7524\\
~ & MemNet~\cite{tai2017memnet} & 2662.4 & 677 & 31.74/0.8893 & 28.26/0.7723 & 27.40/0.7281 & 25.50/0.7630\\
~ & SelNet~(Choi et al. 2017) & 83.1 & 1417 & 32.00/0.8931 & 28.49/0.7783 & 27.44/0.7325 & - \\
~ & CARN~(Ahn et al. 2018) & 90.9 & 1592 & 32.13/0.8937 & 28.60/0.7806 & 27.58/0.7349 & 26.07/0.7837\\
~ & CARN-M~(Ahn et al. 2018) & 32.5 & 412 & 31.92/0.8903 & 28.42/0.7762 & 27.44/0.7304 & 25.62/0.7694\\
~ & ESRN (ours) & 66.1 & 1014 & \textbf{32.26/0.8957} & \textbf{28.63/0.7818} & \textbf{27.62/0.7378} & \textbf{26.24/0.7912}\\
~ & ESRN-F (ours) & 41.4 & 1019 & 32.15/0.8940 & 28.59/0.7804 & 27.59/0.7354 & 26.11/0.7851\\
~ & ESRN-V (ours) & \textbf{20.7} & \textbf{324} & 31.99/0.8919 & 28.49/0.7779 & 27.50/0.7331 & 25.87/0.7782\\
\hline
\end{tabular}}
\end{center}
\end{table*}

We explore and exploit the effectiveness of pooling and recursive operator for super-resolution with experiments. We gradually replace the SRDB with CRDB-O (without recursive) in SRDN (contains 10 SRDBs with 6 convolution layers of 32 channels), and evaluate the performance of each variant model. Figure~\ref{pool_diff} illustrates the ablation results. It shows that a few CRDB-O blocks hardly reduce the PSNR and effectively reduce the Multi-Adds. Thanks to the local residual in block and global concatenation transmitting the original information to the latter block, it alleviates the shortcoming of pooling operator. We also replace the SRDB with the CRDB of recursive 3 to verify the effectiveness of combination of pooling operator and recursive operator. Experimental results in Table~\ref{pool_ablation} and Fig.~\ref{pool_diff} reveal that CRDB effectively improve performance with the constraint of parameter and Multi-Adds. Pooling and recursive operator can effectively expand the receptive field, which is beneficial to predicting the high resolution pixel considering the contextual information. However, when we add too many CRDBs (larger than 5), the PSNR indicator decreases rapidly. 

In addition, we carried out experiments to evaluate the influence of location and arrangement of CRDB and SRDB. The results in Table~\ref{pool_ablation} reveals that first four CRDBs (CRDN-O-F) and cross assignment (CRDN-O-C) achieves better performance than last four CRDBs (CRDN-O-L) assignment. Hence, NAS is adopted to arrange the block type and block parameter of each block for optimal performance. Analyzing the generated networks, we also found that different types of block should cross arrangement. Table \ref{pool_ablation} shows that NAS improves the SR performance with a margin.

\subsection{Analysis of Guided Evolutionary Algorithm}

We compare the training procedures of conventional evolutionary algorithm and the proposed guided evolutionary algorithm for searching efficient SR networks using the same search space and experimental settings. Figure \ref{mutate_diff} illustrates the progress of neural architecture during evolution process, in which each dot denotes the PSNR of an individual elite network in a certain evolution iteration. It reveals that guided mutation effectively accelerates the evolutionary algorithm and finds better network architecture compared with general random mutation. Besides, Table \ref{abl_guided} shows the final performance of searched super-resolution models with different mutation strategy. Guided evolution achieves better performance, it's mainly because the guided mutation induces the algorithm to explore blocks with higher credit preferentially.

\subsection{Comparison with State-of-the-art Methods}

\begin{figure*}[t]
 \vspace{-0.5em}
		\scriptsize
		\centering
		\begin{tabular}{cc}
			\begin{adjustbox}{valign=t}
				\begin{tabular}{c}
					\includegraphics[width=0.35\textwidth]{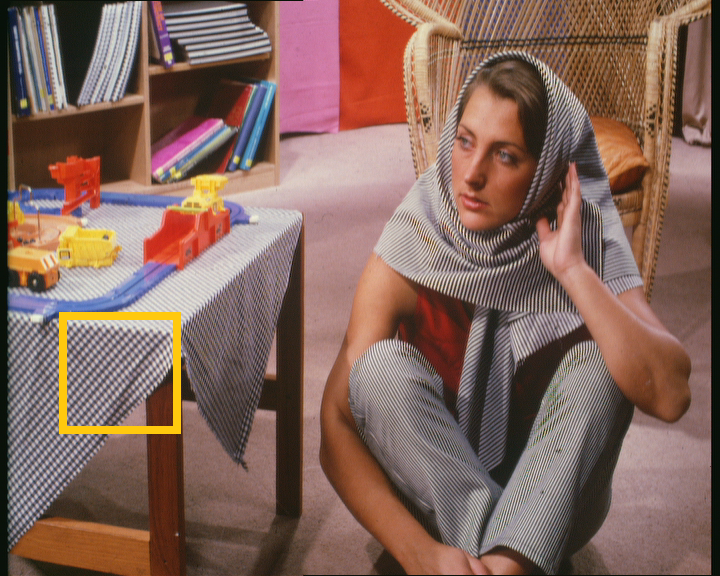}
					\\
					Ground-truth HR
				\end{tabular}
			\end{adjustbox}
			\hspace{-0.32cm}
			\begin{adjustbox}{valign=t}
				\begin{tabular}{cccc}
					\includegraphics[width=0.115\textwidth]{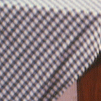} \hspace{-0.25cm} &
					\includegraphics[width=0.115\textwidth]{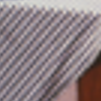} \hspace{-0.25cm} &
					\includegraphics[width=0.115\textwidth]{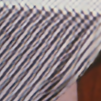} \hspace{-0.25cm} &
					\includegraphics[width=0.115\textwidth]{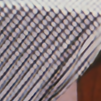}
					\\
					HR\hspace{-0.25cm} &
					Bicubic\hspace{-0.25cm} &
					CARN-M \hspace{-0.25cm} &
					CARN
					\\
					(PSNR, SSIM)\hspace{-0.25cm} &
					(24.01, 0.7460)\hspace{-0.25cm} &
					(21.36, 0.6834)\hspace{-0.25cm} &
					(24.02, 0.8285)
					\\
					\includegraphics[width=0.115\textwidth]{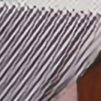} \hspace{-0.25cm} &
					\includegraphics[width=0.115\textwidth]{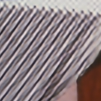} \hspace{-0.25cm} &
					\includegraphics[width=0.115\textwidth]{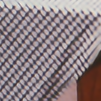} \hspace{-0.25cm} &
					\includegraphics[width=0.115\textwidth]{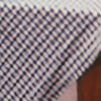} 
					\\ 
					LapSRN \hspace{-0.25cm} &
					VDSR \hspace{-0.25cm} &
					ESRN-V (ours)  &
					ESRN (ours)
					\\ 
					(17.97, 0.4665)\hspace{-0.25cm} &
					(19.80, 0.5637)\hspace{-0.25cm} &
					(26.13, 0.8976)\hspace{-0.25cm} &
					(\textbf{27.42}, \textbf{0.9242}) 
				\end{tabular}
			\end{adjustbox}
		\end{tabular}
 \vspace{-1em}
\caption{Visual qualitative comparison on $\times 3$ scale (barbara from Set14).}
\label{Fig:barbara}
 \vspace{-1.5em}
\end{figure*}

In order to compare the proposed approach with other prominent methods, we search one medium-size super-resolution model, ESRN, and one mini-size super-resolution model, ESRN-V. ESRN contains $1014K$ parameter and ESRN-V contains 324K parameters. These two super-resolution models have the similar number of parameters and Multi-Adds to that of CARN~\cite{ahn2018fast}, SelNet~\cite{choi2017deep} and FALSR~\cite{chu2019fast}. The performance of different models are executed with qualitative and quantitative comparison.

We employ two typical image quality metrics, PSNR and SSIM, to evaluate the performance of super-resolution model. The quantitative comparisons of the performances over the benchmark datasets are summarized in Table~\ref{compare_sota} ($\times 2$ scale) and Table~\ref{scale_sota} ($\times 3$ and $\times 4$ scale). It reveals that the proposed method (ESRN, ESRN-V) achieves better performance than other prominent approaches on all benchmark datasets in the case of the similar amount of parameters and Multi-Adds. In $\times 2$ scale, our medium and mini ESRN models outperform FALSR~\cite{chu2019fast} by a margin of $0.44$ PSNR and $0.51$ PSNR on Urban100 dataset with fewer parameter and Multi-Adds, respectively. The success is mainly due to the proposed  blocks and the guided algorithm.

Different from FALSR \cite{chu2019fast}, relationship of the parameter and Multi-Adds are decoupled by pooling operator. Hence, we can constrain much fewer Multi-Adds to search for a fast medium super-resolution model, ESRN-F. The experimental results in Table \ref{compare_sota} shows that ESRN-F even achieves a little better performance than FALSR-A~\cite{chu2019fast} on all four benchmark datasets with approximate half Multi-Adds. Hence, our evolutionary algorithm could search for a very efficient super-resolution model.

In Fig.~\ref{Fig:barbara}, we visually illustrate the qualitative comparisons on \textit{barbara} of Set14 dataset in $\times 3$ scale. It's obvious that the reconstructed images of other methods contain noticeable artifacts and blurred edges. In contrast, our super-resolution images are more faithful to the ground truth with clear edges. Our ESRN recover image better than other prominent models, since it considers more contextual information with pooling and recursive operator.

\section{Conclusion}
 
In this paper, an efficient residual dense block search algorithm is proposed to hunt for fast, lightweight and accurate super-resolution networks. Three efficient blocks are designed as basic choice components, which reduce the parameter and computation of network in three aspects: channel number, convolution filter and feature scale. With the assistance of evolutionary algorithm, we exploit the variation of feature scale adequately to accelerate super-resolution network. In addition, block credit is introduced to measure the effectiveness of a block and utilized to guide evolution for searching acceleration. Both quantitative and qualitative results demonstrate the advantage of the found network over other state-of-the-art super-resolution methods with specific requirement on number of parameters and FLOPs.

\bibliographystyle{aaai} 
\bibliography{ref}

\end{document}